\documentclass[conference]{IEEEtran}
\IEEEoverridecommandlockouts
\usepackage{cite}
\usepackage{amsmath,amssymb,amsfonts}
\usepackage{algorithmic}
\usepackage{graphicx}
\usepackage{textcomp}
\usepackage{xcolor}
\def\BibTeX{{\rm B\kern-.05em{\sc i\kern-.025em b}\kern-.08em
    T\kern-.1667em\lower.7ex\hbox{E}\kern-.125emX}}
\begin{document}

\title{Using Entropy Measures for Monitoring the Evolution of Activity Patterns\\
}



\author{\IEEEauthorblockN{Yushan Huang}
\IEEEauthorblockA{\textit{Dyson School of Design Engineering} \\
\textit{Imperial College London}\\
\textit{\&}\\
\textit{Care Research and Technology Centre} \\
\textit{The UK Dementia  Research Institute}\\
London, UK \\
yushan.huang21@imperial.ac.uk}
\\
\IEEEauthorblockN{Hamed Haddadi}
\IEEEauthorblockA{\textit{Dyson School of Design Engineering} \\
\textit{Imperial College London}\\
\textit{\&}\\
\textit{Care Research and Technology Centre} \\
\textit{The UK Dementia  Research Institute}\\
London, UK \\
h.haddadi@imperial.ac.uk}
\and
\IEEEauthorblockN{Yuchen Zhao}
\IEEEauthorblockA{\textit{Dyson School of Design Engineering} \\
\textit{Imperial College London}\\
\textit{\&}\\
\textit{Care Research and Technology Centre} \\
\textit{The UK Dementia  Research Institute}\\
London, UK \\
yuchen.zhao19@imperial.ac.uk}
\\
\IEEEauthorblockN{Payam Barnaghi}
\IEEEauthorblockA{\textit{Department of Brain Sciences} \\
\textit{Imperial College London}\\
\textit{\&}\\
\textit{Care Research and Technology Centre} \\
\textit{The UK Dementia  Research Institute}\\
London, UK \\
p.barnaghi@imperial.ac.uk}
}

\maketitle
\begin{abstract}
In this work, we apply information theory inspired methods to quantify changes in daily activity patterns. We use in-home movement monitoring data and show how they can help indicate the occurrence of healthcare-related events. Three different types of entropy measures namely Shannon's entropy, entropy rates for Markov chains, and entropy production rate have been utilised. The measures are evaluated on a large-scale in-home monitoring dataset that has been collected within our dementia care clinical study. The study uses Internet of Things (IoT) enabled solutions for continuous monitoring of in-home activity, sleep, and physiology to develop care and early intervention solutions to support people living with dementia (PLWD) in their own homes. Our main goal is to show the applicability of the entropy measures to time-series activity data analysis and to use the extracted measures as new engineered features that can be fed into inference and analysis models. The results of our experiments show that in most cases the combination of these measures can indicate the occurrence of healthcare-related events. We also find that different participants with the same events may have different measures based on one entropy measure. So using a combination of these measures in an inference model will be more effective than any of the single measures. 
\end{abstract}


\section{Introduction}
 Finding patterns in activity data collected by IoT has been applied to various research fields, including object tracking \cite{ullah2021survey}, intrusion detection \cite{sahoo2017iot}, and healthcare \cite{liu2016human}. Existing research mainly focuses on using machine learning (ML) models and algorithms to learn and analyse patterns from raw data \cite{khan2021deep}. Although such methods can find the direct combination of raw data points that can indicate interesting events, they are not able to utilise statistical and useful information about the data, such as the distribution of data points and the uncertainty in such distributions. Determining the distribution and uncertainty will introduce more useful information into ML models and thus can potentially contribute to building accurate prediction and inference models.
 
\begin{figure}[t]
\vskip 0.2in
\begin{center}
\centerline{\includegraphics[width=200pt]{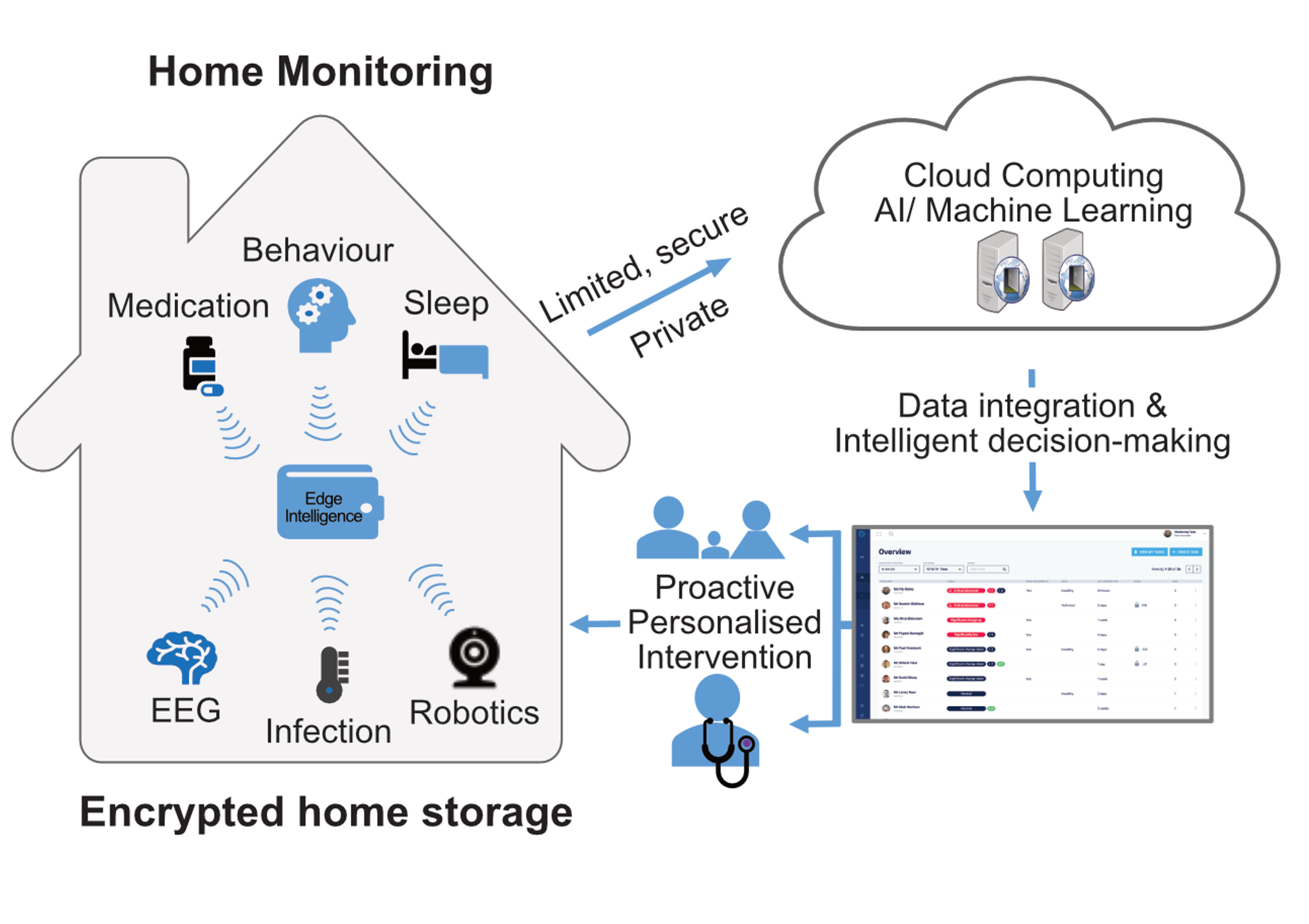}}
\caption{An overview of our in-home IoT monitoring system. The system allows integration of different in-home activity and physiology data. It also provides a framework for deploying and validating analytical models.}
\label{fig:system}
\end{center}
\vskip -0.2in
\end{figure}

 In this paper, we propose three measures that are constructed based on entropy. Our goal is to provide measures to enhance the ability of processing models to quantify the changes in data patterns and improve the outcome of predictive and analytical machine learning models. We empirically evaluated these measures on the data that we have collected in an in-home healthcare monitoring IoT platform (illustrated in Fig.~\ref{fig:system}) to support PLWD. Our preliminary results indicate that, in most cases, combining these new measures into data analysis pipelines can suggest occurrences of certain healthcare-related events. These measures show different suitability on different participants' data. So these measures as engineered  features in modern ML methods can improve the outcomes of inference and predictive models.
 
\begin{table*}[h]
	\centering
	\caption{IoT devices in the platform}
	\label{tab:Iot platform}
	\begin{tabular}{lccr} 
		\hline
		Digital Marker & IoT Device & Frequency \\
		\hline
		Activity & Passive infrared sensors & Triggered by movement\\
		Home device usage & Smart plugs & Triggered by device use\\
		Body temperature & Smart  temporal  thermometers & Twice daily or continuous using a wearable device\\
		Blood pressure and heart rate & Wearable devices & Twice daily\\
		Weight and heart rate & Smart scale with body composition and heart rate & Once a day\\
		Respiratory and heart rate during sleep & Sleep mat & Once a minute\\
		Environmental light & Light sensors & Every 15 minutes\\
		Environmental temperature & Temperature sensors & Once an hour\\
		\hline
	\end{tabular}
\end{table*}

\section{System and data}

We have developed a digital platform, called Minder, to integrate in-home IoT sensors to collect physiological data, sleep data, environmental data, and activity data in a privacy-aware and secure manner. A list of the IoT devices in our platform is shown in Table \ref{tab:Iot platform}. The dataset used in this study includes 9,370 person-day activity data collected between December 2020 and March 2022. We have used the Minder platform to collect remote monitoring data in a dementia study. The study has received ethical approval and all the data used and presented in this research has been anonymised. The Minder platform provides an overview dashboard, which allows a monitoring team to observe raw data and predicted alarms raised by analytical models. The platform has four key components: 1) sensors installed in participants' homes (the platform is designed to be device agnostic), 2) the back-end system including Cloud infrastructure, storage, and analysis tools, 3) the user interface for data visualisation and presenting clinical information, environmental information, and alerts, 4) clinical intervention where healthcare practitioners use the system/alerts and interact with the participants, caregivers and respond to their healthcare needs.

\begin{figure}[t]
\vskip 0.2in
\begin{center}
\centerline{\includegraphics[width=250pt]{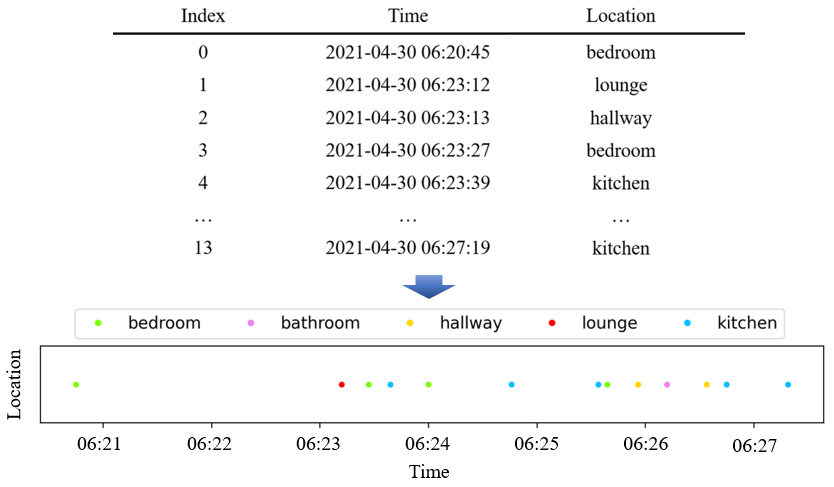}}
\caption{An example for raw Passive Infrared (PIR) data. The data is selected from one participant. The x-axis shows the time of the day, and different colours represent different locations in the house.}
\label{fig:data-vis}
\end{center}
\vskip -0.2in
\end{figure}

\begin{figure*}[h]
\vskip 0.2in
\begin{center}
\centerline{\includegraphics[width=400pt]{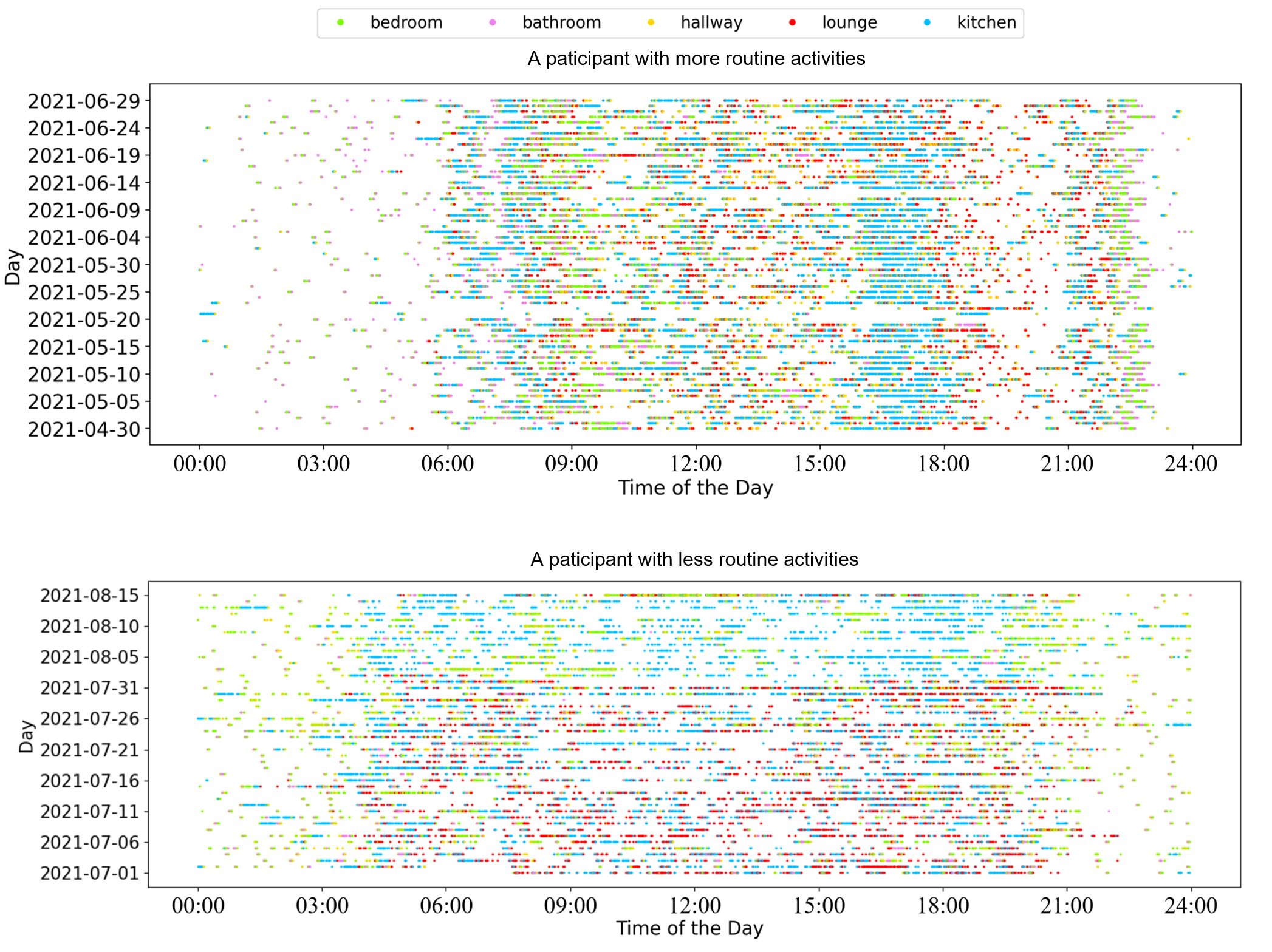}}
\caption{An example of a participant with more routine activities and another participant with less routine activities. The participant with more routine activities tends to have more consistent daily activities at the same time on each day. The x-axis shows the time of the day, the y-axis shows different days, and the different colours represent different locations in the house.}
\label{fig:camparison-routine}
\end{center}
\vskip -0.2in
\end{figure*}

In our system, activity data is collected using passive infrared (PIR) sensors, which are installed in five locations in the home: bathroom, bedroom, lounge, kitchen, and hallway. A PIR sensor in a certain location is triggered when a person passes by and sends an alert to the system at the same time, which records the time and the location of the alert. The raw data is time-series containing location and time information, as Fig.~\ref{fig:data-vis} shows.

The data is labelled by our monitoring team who react to the alerts generated on the platform and verify the alerts by contacting PLWD or their caregivers. The labels contain different healthcare-related events, including accidental falls, abnormal motor function behaviour, hospital admissions, Urinary Tract Infections (UTIs), anxiety and depression, disturbed sleep patterns, agitation, and confusion. We combine the activity data and the labels from different participants to create the dataset.

\section{Methodology}
We aim to capture and model complex features that cannot be directly obtained through linear and nonlinear functions in training models. Fig.~\ref{fig:camparison-routine} shows an example of a participant with more routine activities and a participant with less routine activities. We use entropy to capture the uncertainty from two perspectives including location-based and route-based changes.

\begin{figure*}[h]
\vskip 0.2in
\begin{center}
\centerline{\includegraphics[width=400pt]{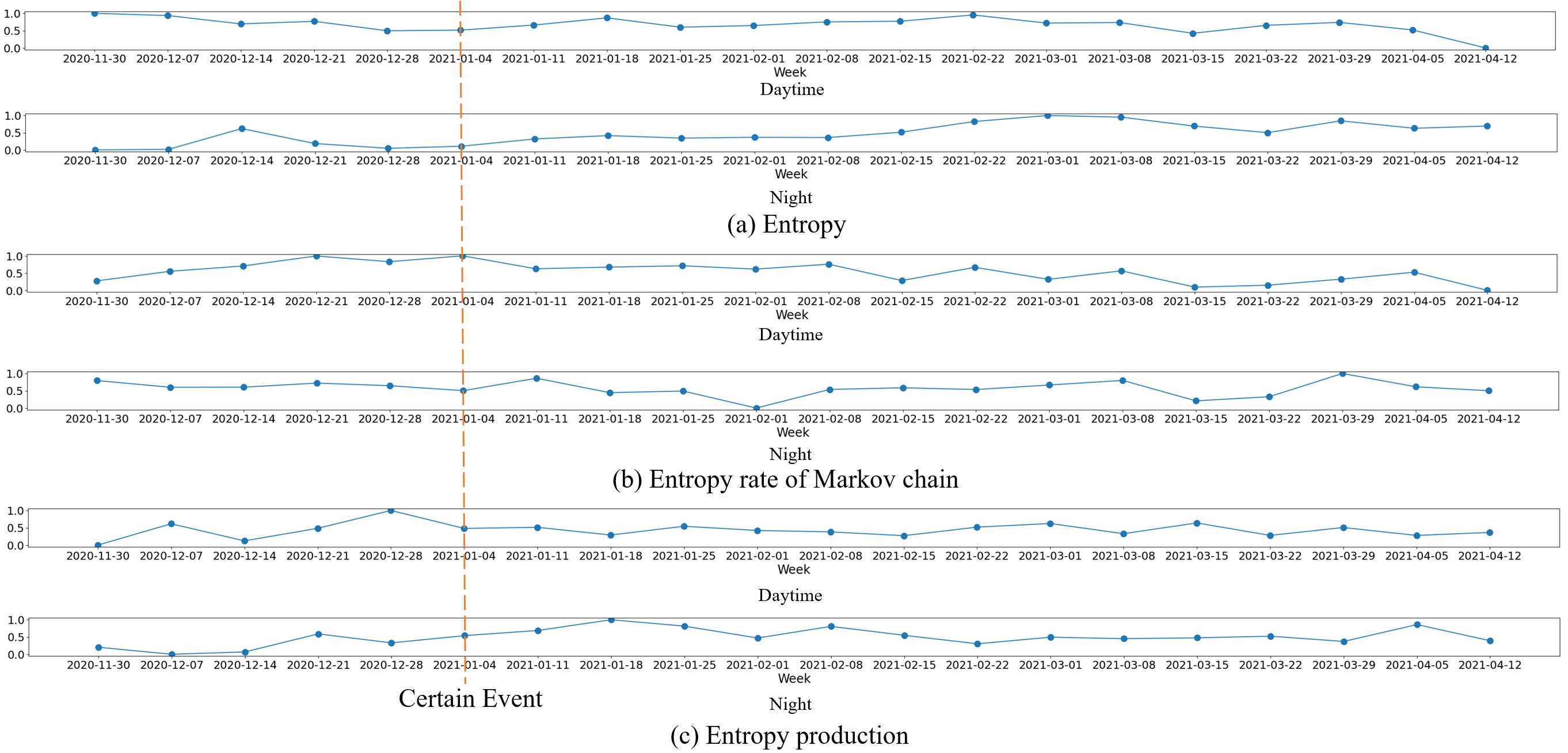}}
\caption{Entropy measures for a household. The x-axis represents Monday of each week, the y-axis represents the average on each day of the week after normalisation. These entropy measures show promising variations and have different performances when the participant was experiencing certain events.}
\label{fig:entropy-compare}
\end{center}
\vskip -0.2in
\end{figure*}

\begin{figure*}[]
\vskip 0.2in
\begin{center}
\centerline{\includegraphics[width=400pt]{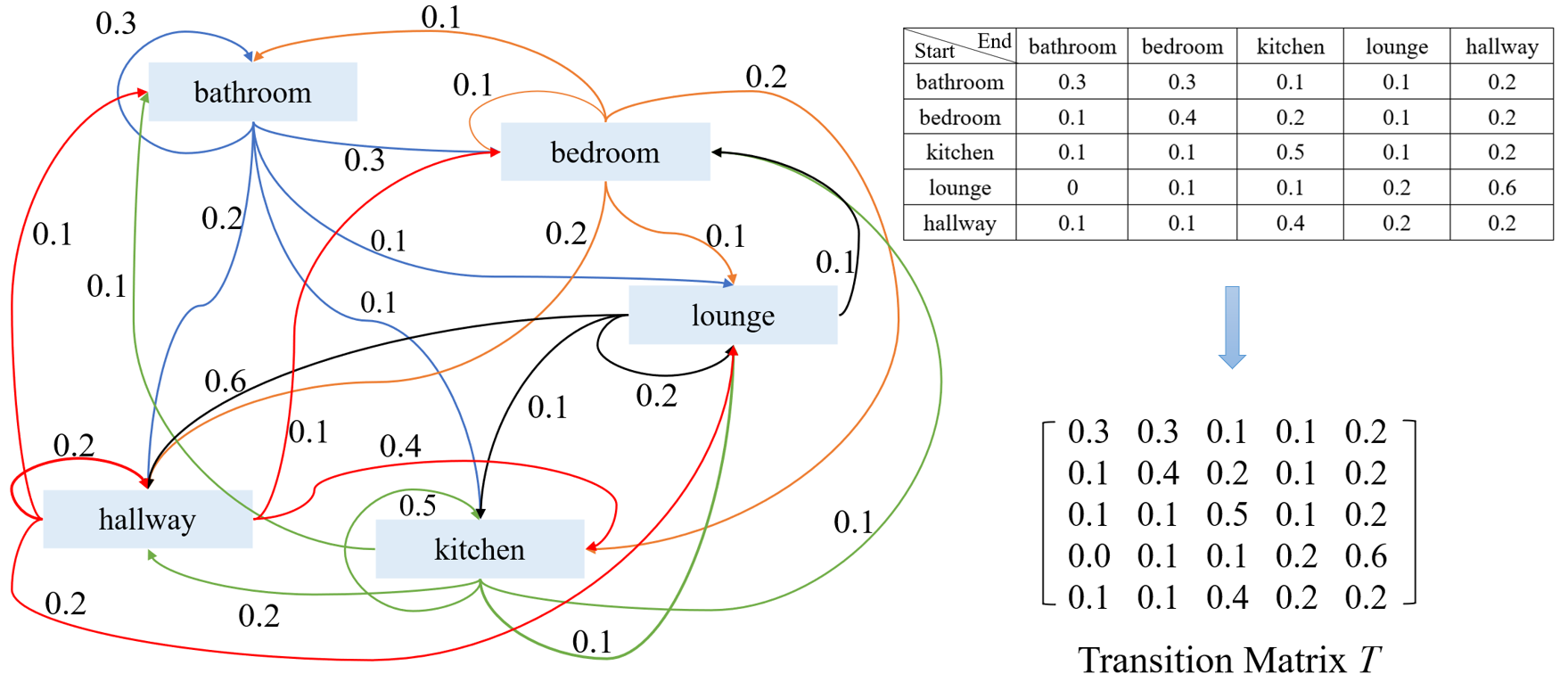}}
\caption{An activity route $X$ with five locations. The rectangular boxes represent the locations, and the arrows represents transition between locations in the house. Different colours represent different start locations (blue: bathroom, orange: bedroom, green: kitchen, black: lounge, and red: hallway). The numbers next to the lines represent transitional probabilities which correspond to the table and Transition Matrix $T$.}
\label{fig:markov}
\end{center}
\vskip -0.2in
\end{figure*}

\begin{figure}[h]
\vskip 0.2in
\begin{center}
\centerline{\includegraphics[width=200pt]{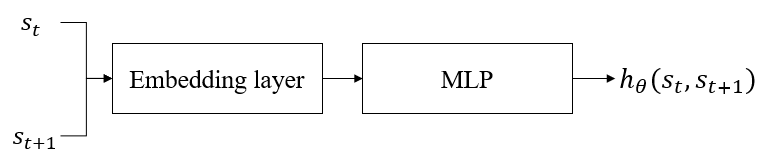}}
\caption{A simplified view of the network to calculate EP.}
\label{fig:network}
\end{center}
\vskip -0.2in
\end{figure}

\subsection{Shannon's entropy}
If the occurrence at locations is regarded as random events, we can consider measuring the extent of occurrence of these random events. Shannon's entropy \cite{shannon1948mathematical} is a conventional method to quantify information. We apply Shannon's entropy to represent changes in activity patterns. Suppose that there are $n$ locations in a participant's activity, denoted as $X = \left\{ x_{1},x_{2},\ldots,x_{n} \right\}$. The Shannon's entropy of the activity data is:

\begin{equation}
    H(X) = - {\sum\limits_{i = 1}^{n}{P\left( x_{i} \right){\log{P\left( x_{i} \right)~}}}}
\end{equation}

In which $P\left( x_{i} \right)$ is the probability at location $x_{i}$. When the activity pattern changes, $H(X)$ will also change accordingly. It shows the change in locations of the participants' activities. Shannon's entropy for the activity data for a sample household is shown in Fig.~\ref{fig:entropy-compare}(a).

\begin{figure*}[th]
\vskip 0.2in
\begin{center}
\centerline{\includegraphics[width=500pt]{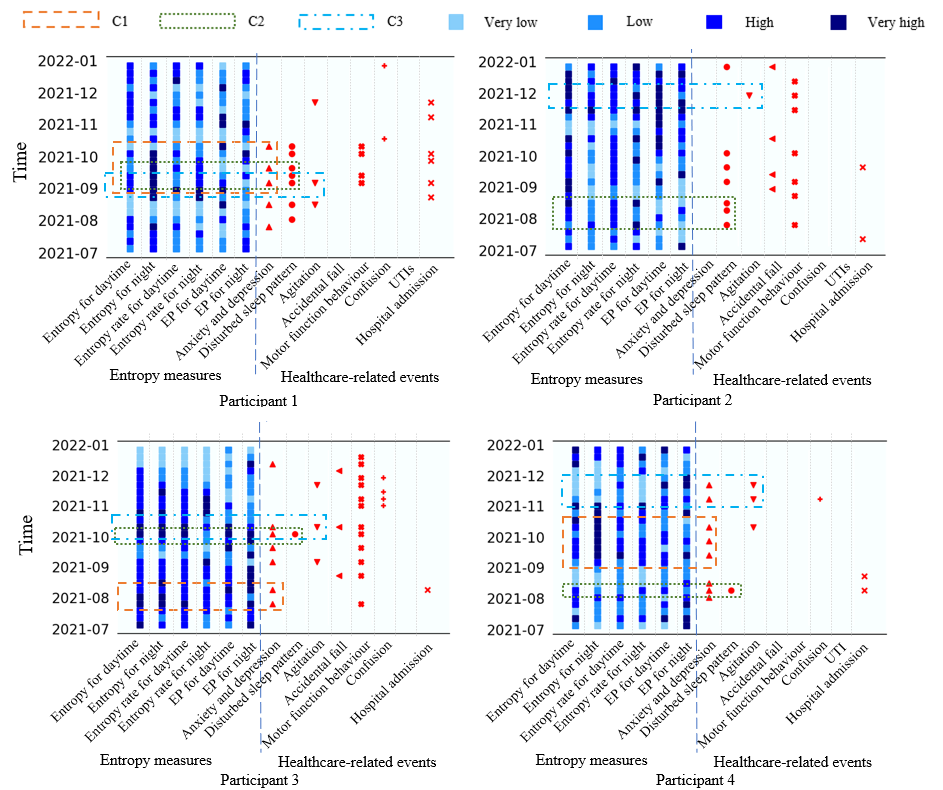}}
\caption{The Entropy results for a group of households. The left part of X-axis represents the Entropy measures and the right part of X-axis represents healthcare-related events. The four varying degrees of blue on the left part represent the four group after the Z-score normalisation. The different shapes in red on the right side represent the participants were experiencing certain healthcare-related events. C1, C2, and C3 are some examples for discussion. C1: the relationship between ``anxiety and depression'' and the entropy measures, C2: the relationship between ``disturbed sleep pattern'' and the entropy measures, C3: the relationship between ``agitation'' and the entropy measures.}

\label{fig:households}
\end{center}
\vskip -0.2in
\end{figure*}

\subsection{Entropy rate of a Markov chain}

If the transition between locations in the home is considered, then the activity data can modelled as a Markov Chain. Compared with healthy individuals, PLWD often have a more non-structured routine \cite{urwyler2017cognitive}. Therefore, finding a method to represent the routineness plays an important role in the identification of participants' healthcare-related events. The first-order Markov chain model is used to profile the living routine of the participants, where the current location of an individual in the home is only dependent on their previous location \cite{tan2008hidden}. Given that $X = \left\{ x_{1},x_{2},\ldots,x_{n} \right\}$ represents $n$ locations in a participant's activity. $l_{a}, l_{b} \in X$, represent the locations at previous moment and current moment. The probability of transitioning from location $l_{a}$ to location $l_{b}$ is: 
 
\begin{equation}
    P_{ij} = P\left( l_{b}=x_{j} \middle| l_{a}=x_{i} \right)
\end{equation}

In which, $x_{i},x_{j} \in X$. The first-order Markov chain model can be represented by $n \times n$ $P_{ij}$ as a Transition Matrix $T$. For example, as shown in Fig.~\ref{fig:markov}, an activity route $X$ with five locations (bathroom, bedroom, kitchen, lounge, and hallway) can be represented by a Transition Matrix $T$. We build the model based on a time window and calculate the entropy rate $\xi$ of the first-order Markov chain model to quantify the changes \cite{enshaeifar2018health}:

\begin{equation}
    \xi = - {\sum\limits_{x_{i},x_{j}\in X}^{n}{P(l_{a}=x_{i})P_{ij}{\log{P_{ij}~}}}}
\end{equation}

We use 16 weeks of activity data to calculate  $T$, and use the rest of the data to calculate the entropy rate. The entropy rate of a sample household is shown in Fig.~\ref{fig:entropy-compare}(b).

\subsection{Entropy production rate}
As another method to describe the transitioning between routes in the home, we use the Entropy Production (EP) rate, which is a description of diverse non-equilibrium principle \cite{dewar2003information}. Some ML models have been proposed to calculate EP, which can be applied to time-series data \cite{kim2022estimating, otsubo2022estimating}. As Fig.~\ref{fig:data-vis} shows, our in-home activity data is time-series data. The in-home movement can be also regarded as a non-equilibrium system \cite{abdelgawwad20193d, marhasev2006non}. Thus, we consider applying an ML method to calculate the EP rate in our activity data. The Neural Estimator for Entropy Production (NEEP) can estimate EP from the time-series data of relevant variables without detailed information \cite{kim2020learning}. 

Given a Markov chain trajectory $S = \left\{ s_{1},s_{2},\ldots,s_{L} \right\}$ and a function $h_{\theta}$ that takes two states $s_{t}$ and $s_{t+1}$, where $\theta$ denotes the trainable neural network parameters, the output of NEEP is defined as:

\begin{equation}
    \mathrm{\Delta}S_{\theta}\left( {s_{t},~s_{t + 1}} \right) \equiv h_{\theta}\left( {s_{t},~s_{t + 1}} \right) - h_{\theta}\left( {s_{t + 1},~s_{t}} \right)
\end{equation}

\begin{equation}
    J(\theta) = E_{t}E_{s_{t}\rightarrow s_{t + 1}}\left\lbrack {\mathrm{\Delta}S_{\theta}\left( {s_{t},~s_{t + 1}} \right) - e^{- \mathrm{\Delta}S_{\theta}{({s_{t},~s_{t + 1}})}}} \right\rbrack
\end{equation}

In which $E_{t}$ denotes the expectation over $t$, which is uniformly sampled from $\left\{ 1,~\ldots,~L - 1 \right\}$, and $E_{s_{t}\rightarrow s_{t + 1}}$ is the expectation over transition $\left. s_{t}\rightarrow s_{t + 1} \right.$. 

In NEEP, an embedding layer is used to transform the discrete state into a trainable continuous vector. The two embedding vectors $s_{t}$ and $s_{t+1}$ are then fed into a hidden multi-layer perceptron (MLP) for $h_{\theta}$ \cite{kim2020learning}. A simplified view of the network is shown in Fig.~\ref{fig:network} . The calculated entropy production rate for a sample household is shown in Figure 4(c).

\subsection{Time windows}
To evaluate the proposed measures, we perform an analysis with a one-week window to investigate the relation between activity pattern changes and healthcare-related events. Choosing a short window such as one week also has the advantage of making our entropy measures not being influenced by seasonal effects. We calculate the three entropy measures on each day of the week, then calculate the average value to represent one week's quantified entropy measures. Due to the effect of the sun-downing and circadian rhythms in PLWD \cite{volicer2001sundowning} and to investigate their potential effect on these measures, we slice one day into different time periods, i.e. daytime (06:00--18:00) and night (18:00--24:00 and 00:00--6:00).

\section{Analysis and Results}
To harmonise the measures across different households' data, we apply Z-score normalisation to each household's data separately, and divide the data equally into four groups based on the normalised entropy values \cite{lin2003symbolic} : very high, high, low, very low. Fig.~\ref{fig:households} demonstrates the results of a group of households. We take C1, C2 and C3 as examples for analysis.

C1 in Fig.~\ref{fig:households} shows the episodes that the participants were experiencing ``anxiety and depression'' (noted by the clinical monitoring team by conducting a weekly survey). For participant 1, the first triangle shows that the EP for daytime became higher, the second and the third triangles show that the entropy for night became higher. For participant 3, the first triangle shows that the EP for daytime and night became higher while the second triangle shows that the entropy for daytime, the entropy for night, the entropy rate for night and EP for night became higher. For participant 4, all the three triangles show that the entropy for night became higher. 

C2 in Fig.~\ref{fig:households} shows that the participants were experiencing ``disturbed sleep pattern''. For participant 1, the entropy for night became significantly higher. For participant 2, the entropy for daytime or the entropy rate for night became higher. In some cases, both of them maybe increased at the same time. For participant 3, almost all the entropy measures became higher except the entropy rate for night. For participant 4, the entropy for daytime, the entropy rate for night and the EP for daytime slightly increased. 

C3 in Fig.~\ref{fig:households} shows that the participants were experiencing ``agitation''. For participant 1, the entropy for night became higher. For participant 2, the entropy for daytime, the entropy rate for night and EP for daytime increased. For participant 3, almost all the entropy measures became significantly higher except the entropy rate for night and the EP for night. For participant 4, the first triangle shows that the EP for daytime and night significantly increased while the second triangle shows that the values of these two entropy measures were slightly higher.

According to our analysis, we find that:

\begin{itemize}
\item In most cases, considering only one entropy method may not give an ideal prediction result, but a combination of these entropy methods can suggest the occurrence of healthcare-related events and might improve the prediction accuracy. For example, in C1 of Fig.~\ref{fig:households}, when participant 1 experiencing ``anxiety and depression'', if we only consider the entropy for night, it is difficult to predict the case which is represented by the first triangle.

\item The same healthcare-related events may be associated with different entropy measures in various cases. For example, in C2 of Fig.~\ref{fig:households}, when experiencing ``anxiety and depression'', almost all the values of the entropy measures of participant 3 are higher. However, for participant 4, only the EP for daytime and night show significant changes, while the other entropy measures stay normal. C1 and C3 in Fig.~\ref{fig:households} also show similar results.
\end{itemize}

\section{Conclusion}
In this paper, we have proposed three different types of entropy measures that cannot be directly constructed through training ML models. Based on our preliminary results, we identify a number of directions for future research. Our results show that our proposed measures can be used to indicate activity changes using in-home movement monitoring data. The activity changes identified by changes in entropy measures can then be used to identify the occurrence or risk of healthcare-related events. The retrospective analysis presented in this paper is based on historical data. We plan to deploy these measures in our digital platform and evaluate the clinical utility of these measures in real-world settings in our current dementia care study. 
\section*{Acknowledgment}

This project is supported by the EPSRC PROTECT Project (grant number: EP/W031892/1) and the UK DRI Care Research and Technology Centre funded by MRC and Alzheimer’s Society (grant number: UKDRI-7002). Yushan Huang is funded by China Scholarship Council.

\bibliography{sample}
\bibliographystyle{unsrt}

\end{document}